\documentclass[letterpaper]{article} 
\usepackage[submission]{aaai25}  
\usepackage{times}  
\usepackage{helvet}  
\usepackage{courier}  
\usepackage[hyphens]{url}  
\usepackage{graphicx} 
\usepackage{natbib}  
\usepackage{caption} 
\usepackage{algorithm}
\usepackage{algorithmic}
\usepackage{amssymb}
\usepackage{amsmath}
\usepackage{multirow}
\usepackage{marginnote}  
\usepackage{newfloat}
\usepackage{listings}
\DeclareCaptionStyle{ruled}{labelfont=normalfont,labelsep=colon,strut=off} 
\floatstyle{ruled}
\newfloat{listing}{tb}{lst}{}
\floatname{listing}{Listing}
\title{TG-LLaVA: Text Guided LLaVA via Learnable Latent Embeddings\thanks{Work was done when D. Yan was an intern at Alibaba}}
\author{
    Dawei Yan\textsuperscript{\rm 1,2},
    Pengcheng Li\textsuperscript{\rm 2},
    Yang Li\textsuperscript{\rm 2},
    Hao Chen\textsuperscript{\rm 3},
    Qingguo Chen\textsuperscript{\rm 2},
    Weihua Luo\textsuperscript{\rm 2}, \\
    Wei Dong\textsuperscript{\rm 1},
    Qingsen Yan\textsuperscript{\rm 1},
    Haokui Zhang\textsuperscript{\rm 1\footnote{Corresponding author}},
    Chunhua Shen\textsuperscript{\rm 3}
}
\affiliations{
    \textsuperscript{\rm 1}Northwestern Polytechnical University\\
    \textsuperscript{\rm 2}AI Business, Alibaba Group\\
    \textsuperscript{\rm 3}Zhejiang University
    
}

\usepackage{bibentry}

\begin{document}

\maketitle

\begin{abstract}
Currently, inspired by the success of vision-language models (VLMs), an increasing number of researchers are focusing on improving VLMs and have achieved promising results. However, most existing methods concentrate on optimizing the connector and enhancing the language model component, while neglecting improvements to the vision encoder itself. In contrast, we propose Text Guided LLaVA (TG-LLaVA) in this paper, which optimizes VLMs by guiding the vision encoder with text, offering a new and orthogonal optimization direction. Specifically, inspired by the purpose-driven logic inherent in human behavior, we use learnable latent embeddings as a bridge to analyze textual instruction and add the analysis results to the vision encoder as guidance, refining it. Subsequently, another set of latent embeddings extracts additional detailed text-guided information from high-resolution local patches as auxiliary information. Finally, with the guidance of text, the vision encoder can extract text-related features, similar to how humans focus on the most relevant parts of an image when considering a question. This results in generating better answers. Experiments on various datasets validate the effectiveness of the proposed method. 
Remarkably, without the need for additional training data, our propsoed method can bring more benefits to the baseline (LLaVA-1.5) compared with other concurrent methods. Furthermore, the proposed method consistently brings improvement in different settings. 
Code will be made available upon publication.
\end{abstract}

%

\section{Introduction}
By incorporating visual information into large language models (LLMs), visual language models (VLMs) build on the success of LLMs like ChatGPT~\cite{chatgpt} and Llama~\cite{touvron2023llama}, taking their capabilities a step further. VLMs are not limited to language-based dialogue with humans, they can also discuss the image content, answer questions related to the visual inputs, etc. Recently, centered around VLMs, researchers have conducted extensive work and have made significant progress~\cite{wu2023gpt4vis,zhang2024mm,awadalla2023openflamingo,reid2024gemini}. 

\begin{figure}[t]
\centering
\includegraphics[width=0.45\textwidth]{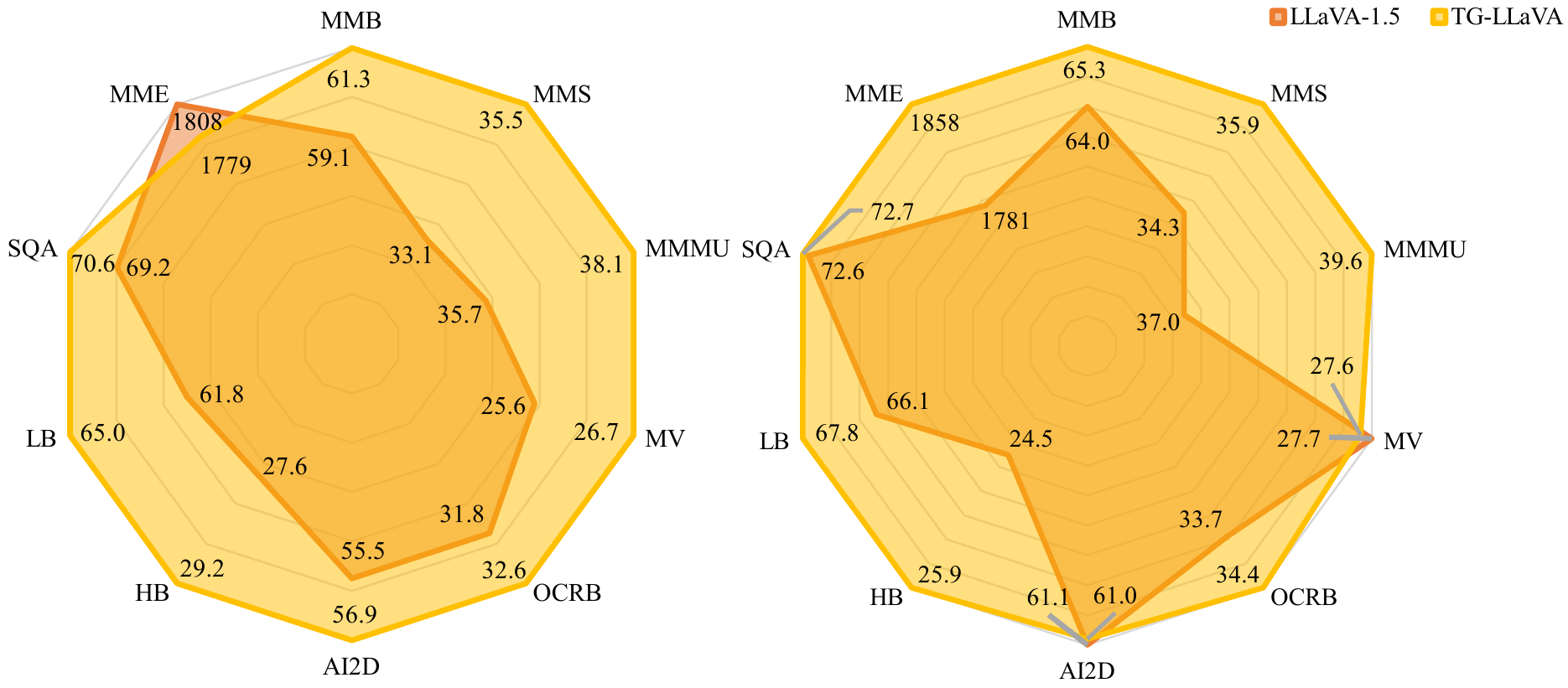} 
\caption{Percentage performance improvements of TG-LLaVA over the baseline LLaVA-1.5~\cite{liu2024improvedbaselinesvisualinstruction} across ten benchmarks, using Vicuna-7B (left) and 13B (right), respectively.}
\label{reda}
\end{figure}

Current adopted VLMs typically consist of three main components: vision encoder, large language model, and connector. The vision encoder, trained on vast amounts of image-text pairs using contrastive learning, encodes images into a shared space with text. Widely used examples include CLIP~\cite{radford2021learning} and SigLIP~\cite{zhai2023sigmoid}. LLMs such as Llama~\cite{touvron2023llama}, Vicuna~\cite{chiang2023vicuna}, Qwen~\cite{bai2023qwen}, and Yi~\cite{young2024yi} have made significant strides in natural language processing tasks, paving the way for integrating vision with text in VLMs. Connector focuses on aligning visual and language features, serving as bridges between modalities. 

Corresponding to the main architecture of VLMs, current improvement methods primarily focus on optimizing the connector and enhancing the language model component among the three major components. For instance, BLIP2~\cite{li2023blip} carefully design multiple loss functions for both contrastive and generative learning, which allows it to achieve precise cross-modal alignment through a multi-stage training process. MoE-LLaVA~\cite{lin2024moe} incorporates a mixture of experts into the second feature forward network layer to enhance the connector component. DenseConnector~\cite{yao2024dense} uses dense connections to merge features from various levels, providing more visual information to the LLM. ImageBind-LLM~\cite{han2023imagebind} transforms image features using a binding network and then integrates these transformed features with the word tokens of the LLM. Besides improving the model structure, increasing the amount of data is a commonly employed strategy. This method usually yields more noticeable results, but it also involves a significantly greater workload. 

In this paper, we propose Text-Guided LLaVA (TG-LLaVA), which optimizes the Visual Language Model from a different even contrasting perspective. Unlike previous work that focuses on enhancing the connector or LLM components, our approach concentrates on improving the visual encoder itself. In contrast to the main strategy that integrating image features into the LLM, we integrate text-guided information into the image features. 

The basic idea of our TG-LLaVA is motivated by two key insights: 1) When humans solve visual question answering tasks, they use the question as a prior, selectively focusing on local regions or specific targets to observe and respond. 2) Numerous studies have demonstrated that improved visual representations are crucial for enhancing VLM performance. The proposed TG-LLaVA aims at guiding the visual encoding process of current VLMs using textual instructions, thereby optimizing the visual branch of VLMs. Specifically, the proposed TG-LLaVA contains two text guided modules, text-guided feature optimization mask (TG-FOM) module and text-guided and text-guided detail perceiver (TG-DP) module. In TG-FOM module, a set of learnable latent embeddings is used to analyze the input text from the global view, then the analyzed language information is added to image feature via a zero-initialized linear layer as guidance. In TG-DP module, a very small number of learnable latent embeddings are used to parse the input text in detail, then the parsed tokens are used as guidance to fuse information from focused image perspective. 
As shown in Figure~\ref{reda}, extensive experiments have demonstrated the effectiveness of the proposed design, showing significant improvements over the baseline across multiple datasets and different framework without the need for any additional data augmentation or complex enhancements. 
Main contributions are summarized as follows:

\begin{itemize}

\item We propose TG-LLaVA, a text-guided architecture based on learnable latent embeddings which is different even opposite to most of existing VLM optimization approaches, which open up a new and worthwhile research avenue for consideration. 

\item The proposed TG-FOM module and TG-DP module can be universally applied as a modular plugin to mainstream VLM frameworks, consistently brings improvement. 

\item Through extensive experiments on various settings of VLM variations and numerous multimodal tasks, we show that our proposed TG-LLaVA not only delivers substantial benefits but also provides valuable insights and methodologies for the existing VLM research field.


\end{itemize}

\section{Related Work}
\subsection{Vision Language Models}
Visual language models primarily consist of a visual encoder and a large language model, representing prominent architectures in the multimodal domain. Researchers have proposed numerous architectures~\cite{li2023mimic,zhu2024ibd,chen2023pali} for integrating visual features into advanced LLM inference pipelines. Llama-Adapter~\cite{zhang2023llama} proposes to generate language answer with taking the image input as condition. Flamingo~\cite{alayrac2022flamingo} and LLaVA~\cite{liu2024visual} blend visual tokens with text as inputs to LLM, differing in that Flamingo employs gating mechanisms to inject encoded visual features into LLMs, while LLaVA directly concatenates visual and textual features at input. Complementarily, the availability of high-quality image-text pairs for VLM training is crucial. Several methods use Chat-GPT~\cite{chatgpt} and GPT-4~\cite{gpt4V} to construct large-scale, high-quality datasets~\cite{zhu2023minigpt,liu2024visual,zhao2023svit}. 

Inspired by the compact structure and outstanding performance of LLaVA-1.5~\cite{liu2024improvedbaselinesvisualinstruction}, we use LLaVA-1.5 as our baseline and incorporate a text-guided approach, similar to other LLaVA-based methods. Unlike most of these methods, which create additional datasets to enhance performance, our improvements focus entirely on the model architecture itself. This approach can further enhance the performance of methods that rely on extra datasets. The results in fifth and sixth lines of Table~\ref{compare_baseline} verified this point. 


%



\begin{figure*}[ht]
\centering
\includegraphics[width=0.7\textwidth]{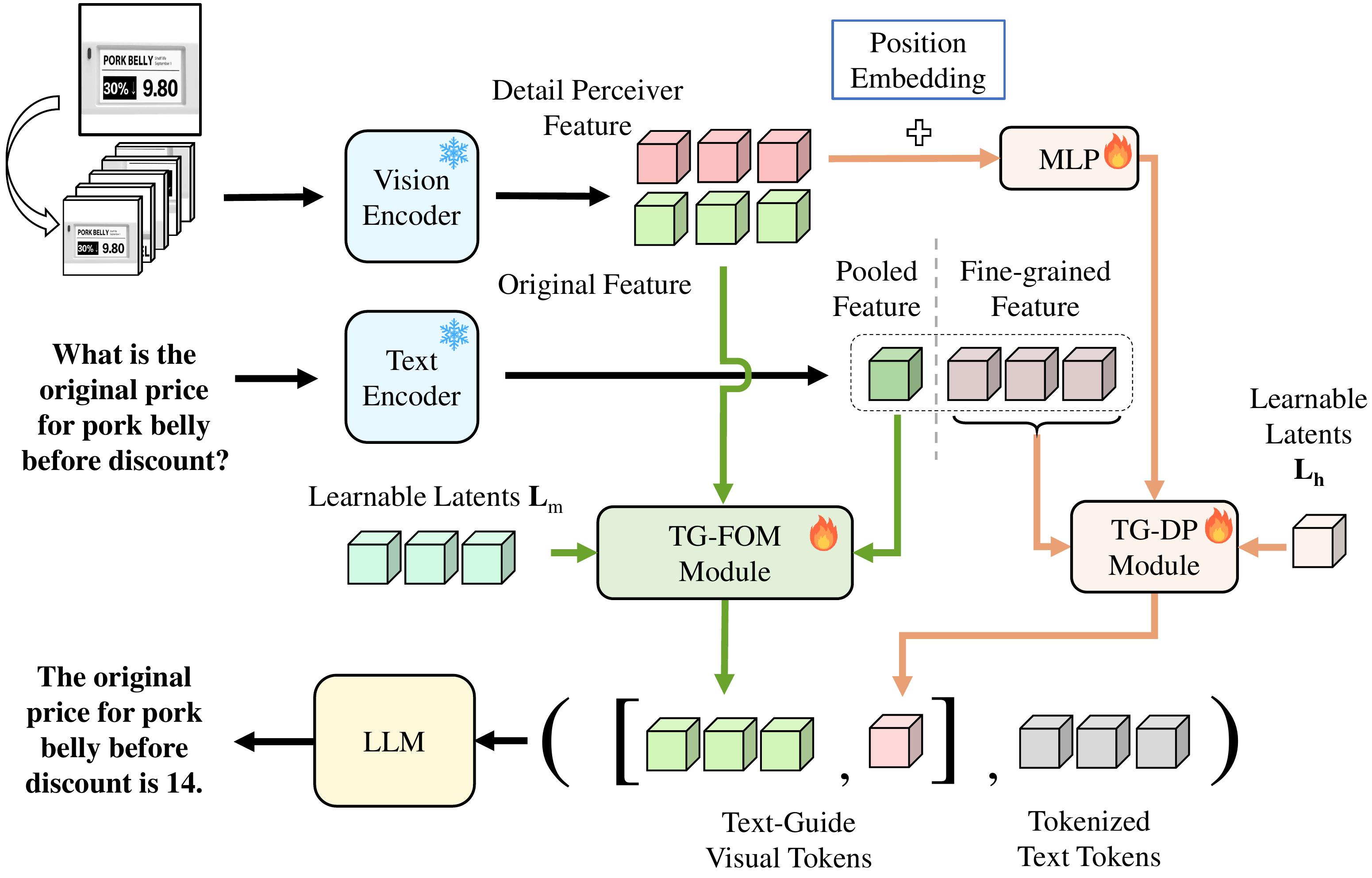} 
\caption{Overall framework of the proposed TG-LLaVA. Text-guided visual feature optimization mask (TG-FOM) module is designed to optimize the visual feature with the guidance of global text. Text-guided detail perceiver (TG-DP) module is proposed to capture instruction relevant details.}
\label{fig:overall_framework}
\end{figure*}

\subsection{Image-Text Alignment}
Align the visual and text information in high semantic level is the base for building VLMs. Centered around this problem, researchers have done extensive work. Previous researchers have typically employed contrastive learning across modalities and autoregressive learning for text. CLIP~\cite{radford2021learning} and SigLIP~\cite{zhai2023sigmoid} trained encoders on massive datasets, laying foundational work for aligning visual and textual modalities and significantly advancing subsequent VLM developments. BLIP~\cite{li2022blip} meticulously design multiple loss functions for contrastive and generative learning, achieving refined cross-modal alignment through multi-stage training. BLIP-2~\cite{li2023blip} adopts a Q-former structure, interacting with the visual modality using learnable query vectors before merging with the text modality. Many LLaVA-like approaches use simple MLPs for modal alignment, with subsequent works like MobileVLM V2~\cite{chu2024mobilevlm}. 

Both image-text alignment methods and our proposed TG-LLaVA recognize the importance of integrating textual and visual information. However, while these methods focus on bridging different modalities, our approach leverages the textual modality to guide and optimize the visual modality. This alignment makes the operation of VLMs more consistent with the purpose-driven logic of human behavior in real-world scenarios.


%

%


\subsection{Visual encoder in VLMs}
To enable the LLM to extract more information from the input visual image, various strategies have been proposed for utilizing visual features. DenseConnector~\cite{yao2024dense} employs dense connections to link visual features across different levels, feeding the combined features into a connector. TokenPacker~\cite{li2024tokenpacker} merges visual features from the high-resolution branch with those from the low-resolution branch to generate condensed visual tokens. Idefics2~\cite{laurenccon2024matters} compresses visual features using a perceiver structure, significantly reducing the number of visual tokens compared to other approaches. Approaches like Mini-Gemini~\cite{li2024mini}, LLaVA-Next~\cite{liu2024llava}, Qwen-VL~\cite{bai2023qwenvl}, and InterVLM~\cite{dong2024internlm} leverage high-resolution images to capture finer visual feature details. ImageBind-LLM~\cite{han2023imagebind} and Llama3.1~\cite{llama31} explore injecting visual modality features into LLMs, with the former using trainable gating modules to add visual features to word tokens, and the latter introducing visual information across different layers of LLM through periodic cross-attention. 

Unlike methods that focus on better utilizing existing visual features, our proposed TG-LLaVA aims to enhance the visual features themselves by using textual guidance. In contrast to ImageBind-LLM and Llama3.1, which incorporate image features into the LLM component, our approach integrates text into the visual encoder.

\section{Method}

In this section, we first review the classic VLM architecture, using LLaVA~\cite{liu2024visual} as a representative example, to provide an overview of the VLM paradigm. Following this, we present a detailed explanation of the proposed TG-LLaVA architecture, focusing on the implementation of the two text-guided modules, text-guided visual feature optimization mask module and text-guided detail perceiver module. 



\subsection{A Revisit of VLMs}

Taking LLaVA~\cite{liu2024visual} as an example, the primary goal of VLMs is to effectively harness the capabilities of pre-trained LLM and visual model. The three key components of such framework can be defined as follows: 

1) \textbf{Visual encoder} \({\rm E}_{\rm v}\), typically utilizing a pre-trained vision transformer like CLIP, is designed to partition the input image \(\mathbf{I} \in \mathbb{R}^{H\times W \times C}\) into several patches with equal size and further encode them into visual features \(\mathbf{F}_{\rm i} \in \mathbb{R}^{N \times D}\). Here, \(H\) and \(W\) represent the size of the input image, \(C\) denotes the number of channels, \(N\) corresponds to the number of patches in the output features, and \(D\) represents the feature dimension of each encoded patch. When the patch size is \(P\), \(N =H W / P^2 \). 2) \textbf{Connector} \({\rm C}\) (also referred as Projector) consists of two linear layers with a GELU activation function in between. Its purpose is to map visual features into the embedding space of the LLM, converting \(\mathbf{F}_{\rm i}\) into visual tokens \(\mathbf{T}_{\rm v}\).
3) \textbf{LLM} \({\rm L}\) employs a tokenizer and text embedding module to sequentially transform textual data into token IDs and their corresponding embedded tokens \(\mathbf{T}_{\rm t}\), effectively converting the language into the feature space of its input. Within the VLM architecture, these textual tokens \(\mathbf{T}_{\rm t}\) are concatenated with the aligned visual tokens \(\mathbf{T}_{\rm v}\) processed by the connector, forming the input for the LLM to carry out subsequent predictions.
For a sequence of length \(L\), the probability of VLM predicting the target answer tokens \(\mathbf{T}_{\rm a}=\left\{t_i\right\}_{i=1}^L\) can be formalized as:

\begin{equation}
p\left(\mathbf{T}_{\rm a} \mid \mathbf{T}_{\rm v}, \mathbf{T}_{\rm t}\right)=\prod_{i=1}^L p_{\theta}\left(t_i \mid \mathbf{T}_{\rm v}, \mathbf{T}_{{\rm t},<i}, \mathbf{T}_{{\rm a},<i}\right),
\end{equation}
where \(\theta\) represents all the trainable parameters in the VLM. In this VLM prediction paradigm, the visual features are directly obtained by encoding the raw input image through \({\rm E}_{\rm v}\) without any interaction with the textual modality. This approach contrasts with the purpose-driven nature of human behavior. Optimizing the encoded features based on textual instructions is more conducive to enabling the VLM to generate accurate responses.

\begin{figure}[t]
\centering
\includegraphics[width=0.32\textwidth]{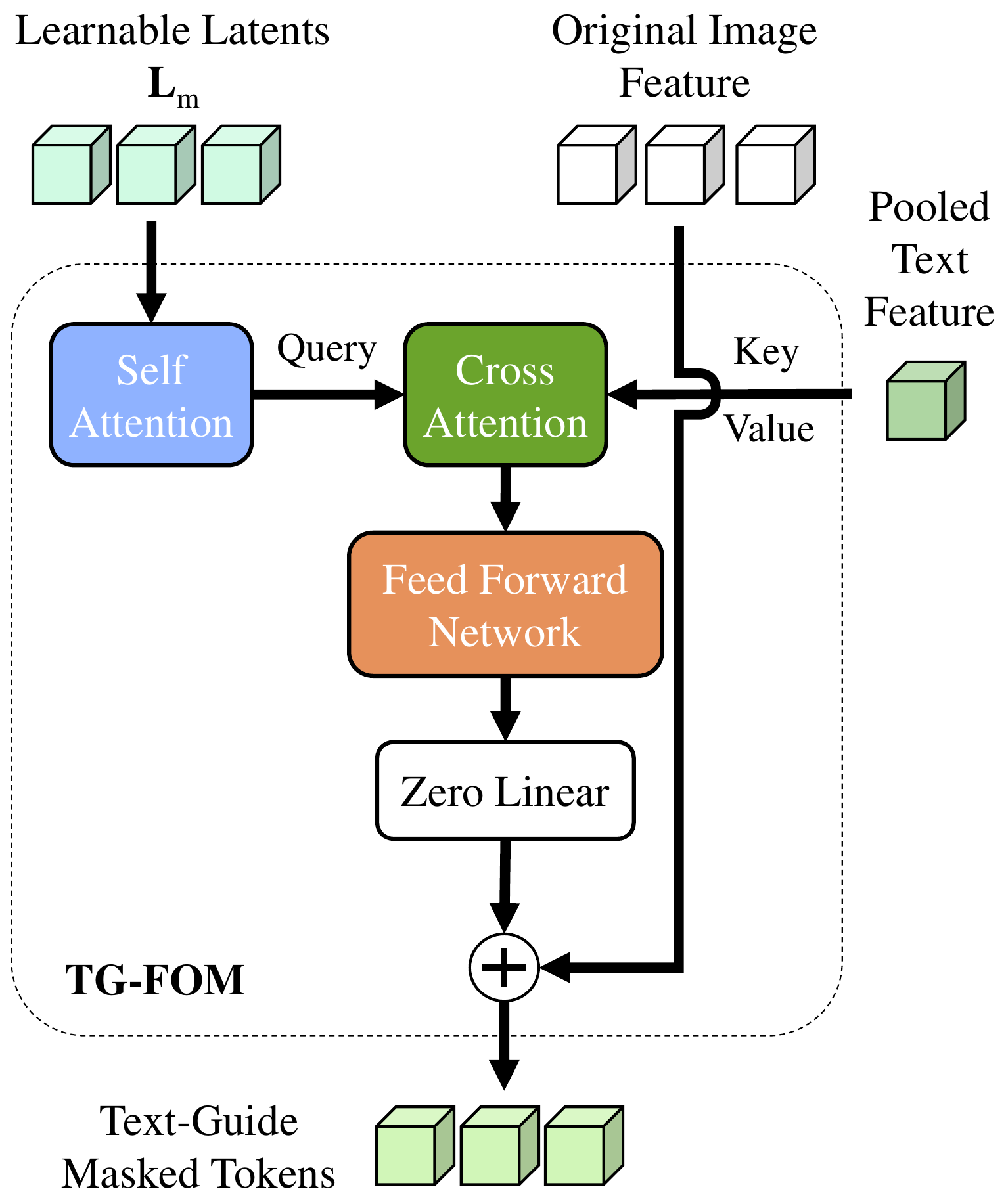} 
\caption{Illustration of Text-guided Visual Feature Optimization Mask module.}
\label{fig:TG_FOM}
\end{figure}

\subsection{Text Guided LLaVA}
Inspired by the reasoning logic humans use in visual question answering scenarios, we design TG-LLaVA, a novel approach that optimizes visual features to align the inference process of VLM more closely with purpose-driven human behavior, thereby further enhancing the capabilities of VLMs. As illustrated in Figure~\ref{fig:overall_framework}, TG-LLaVA primarily consists of two components: Text-guided Visual Feature Optimization Mask (TG-FOM) and Text-guided Detail Perceiver (TG-DP). The former uses learnable latents to parse the global information from textual instructions and attaches it as a mask to the output of visual encoder, optimizing features based on textual instructions. The latter employs another set of latents, first interacting with the detailed information from textual instructions, and then extracting fine-grained details from high-resolution patches of the input image based on these instructions. These details are concatenated with the original features, further refining the visual modality input of VLM. The specifics of this approach will be elaborated in the following sections.

\begin{figure}[t]
\centering
\includegraphics[width=0.45\textwidth]{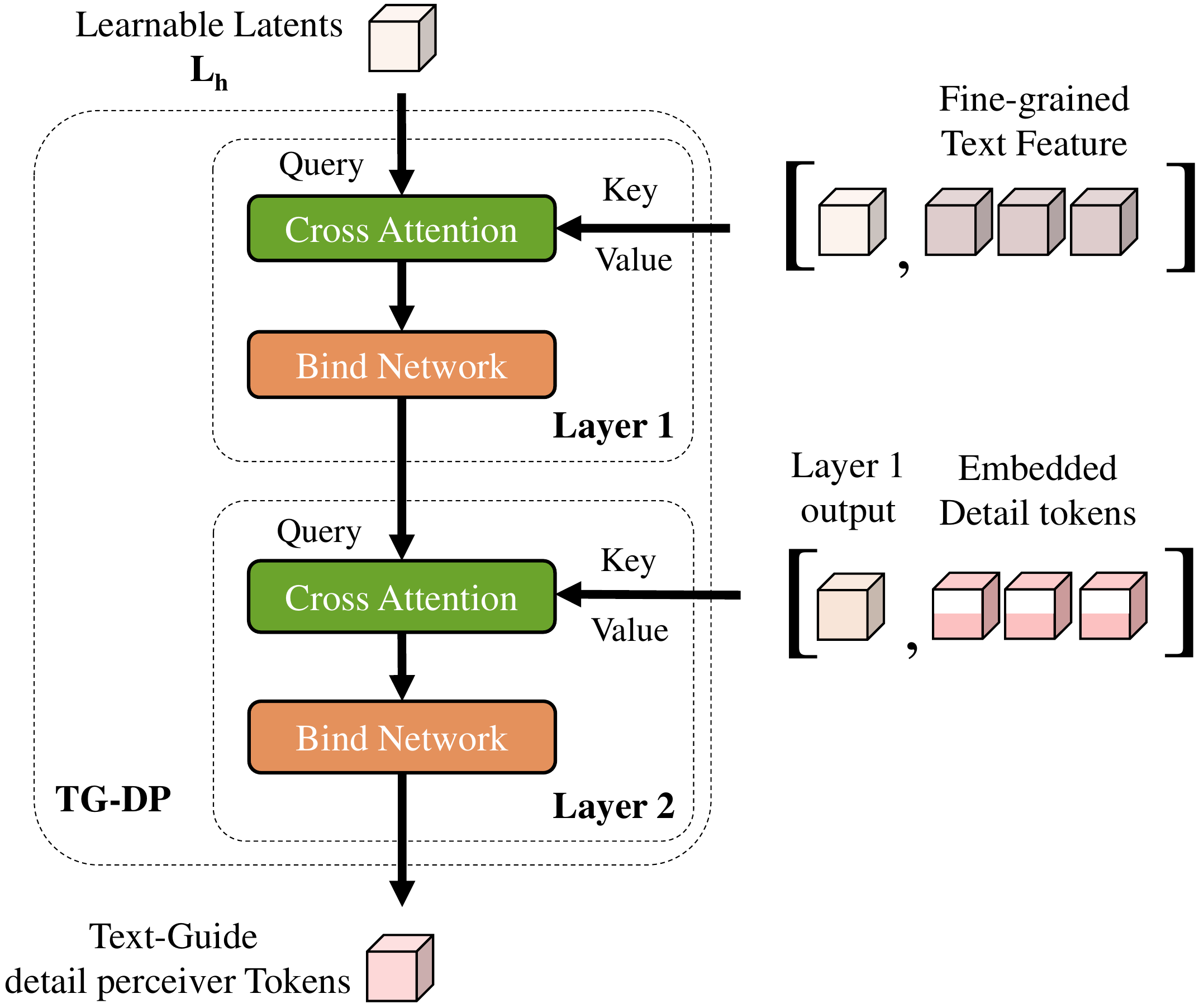} 
\caption{Illustration of Text-guided Detail Perceiver module.}
\label{fig:TG_DP}
\end{figure}

\subsubsection{Text-Guided Visual Feature Optimization Mask}
In current VLMs, the visual representations typically originate solely from the final layer features of the visual encoder \({\rm E}_{\rm v}\). Features obtained through this pipeline encompass the global information of the input image \(\mathbf{I}\). However, the corresponding textual instructions often focus on specific local targets within the image. As a result, the information related to these focal targets is easily compromised when confronted with irrelevant or even contradictory information, leading to distorted judgments by the VLM. To address this issue, we design TG-FOM module to optimize visual features based on textual instructions, thereby endowing VLMs with the advantage of purpose-driven human behavior. Figure \ref{fig:TG_FOM} illustrates the specific framework of the FOM module.

We begin by initializing a set of learnable latent embeddings \(\mathbf{L_{\rm m}}\) that are of the same number as visual tokens. The purpose of these latents is to extract linguistic information from the textual instructions and add it as a mask to the original features. Here, we design a single-layer Q-former to parse semantic information from textual instructions, serving as a bridge between global text and visual features. In this structure, the cross-attention layers incorporate the pooled textual instruction features \(\mathbf{F}_{\rm t}^{\rm p}\) encoded by CLIP text encoder \({\rm E}_{\rm t}\) as Key and Value for interaction with Query \(\mathbf{L_{\rm m}}\), and the final output is a mask generated based on the textual information, which is then applied to the visual features. We additionally introduce a zero-initialized linear layer to ensure that the optimization of the original visual features remains a gradual process. This process can be formalized as:

\begin{equation}
\begin{split}
\mathbf{M}_{\rm t}={\rm Q}(\mathbf{L}_{\rm m}, \mathbf{F}_{\rm t}^{\rm p})&
={\rm FFN({{\rm A}_{\rm cross}({{\rm A}_{\rm self}(\mathbf{L}_{\rm m})}, \mathbf{F}_{\rm t}^{\rm p})})},\\
\mathbf{F}_{\rm i}^*&=\mathbf{F}_{\rm i} + {\rm Z}(\mathbf{M}_{\rm t}),
\end{split}
\end{equation}
where \(\mathbf{M}_{\rm t}\) represents the mask obtained by extracting semantic information from the textual instructions via the learnable \(\mathbf{L_{\rm m}}\), \({\rm A}_{\rm cross}\) and \({\rm A}_{\rm self}\) denote the cross-attention and self-attention modules, respectively, \({\rm FFN}\) represents the feed-forward neural network, \({\rm Z}\) represents the zero-initialized linear layer used as a buffer during feature addition, and \(\mathbf{F}_{\rm i}^*\) represents the visual features optimized by text guidance.


\subsubsection{Text-Guided Detail Perceiver}
When observing images, in addition to selecting focal points based on the instruction, humans can also automatically adjust their focus to obtain more detailed information. Following this idea, we design TG-DP, which is responsible for capturing instruction relevant details. 

As shown in Figure~\ref{fig:overall_framework}, we scale up the original image \(\mathbf{I}\) to preserve more details, then divide it into patches that match the size of the original image. This design ensures that we can extract all visual features with a single call to the visual encoder. After obtaining the visual features of these patches, we add positional embeddings and a learnable MLP layer to recover the spatial structure information that was disrupted during the division operation, getting corrected visual features \(\mathbf{F}_{\rm i}^{\rm h}\). So far, the visual tokens containing detailed information are ready. Along with the learnable latent embeddings \(\mathbf{L}_{\rm h}\) and the fine-grained textual instruction features \(\mathbf{F}_{\rm t}^{\rm g}\), these visual tokens will be fed inoto the TG-DP module, where they will be selected and integrated according to the guidance of the text.

As shown in Figure~\ref{fig:TG_DP}, we set up \(\mathbf{L}_{\rm h}\) to interact with \(\mathbf{F}_{\rm t}^{\rm g}\) output by the text encoder \({\rm E}_{\rm t}\). Here, the number of \(\mathbf{L}_{\rm h}\) is much smaller than the number of the original visual tokens, ensuring that the visual tokens input to the LLM do not increase significantly, thus maintaining inference efficiency. Ablation studies demonstrate that this compression does not negatively impact the final results. 

The proposed TG-DP module consists of two perception layers: 
\begin{itemize}
\item The first perception layer is responsible for parsing fine-grained text to generate text guidance tokens. It receives \(\mathbf{L}_{\rm h}\) and \(\mathbf{F}_{\rm t}^{\rm g}\), maintaining \(\mathbf{L}_{\rm h}\) as the Query and \(\mathbf{F}_{\rm t}^{\rm g}\) as the Key and Value, with the distinction that \(\mathbf{F}_{\rm t}^{\rm g}\) is concatenated with \(\mathbf{L}_{\rm h}\).

\item The second layer is in charge of generating detail perceiver tokens with the guidance of fine-grained text. In the second layer, the Key and Value are replaced by \(\mathbf{F}_{\rm i}^{\rm h}\), using textual instruction features parsed through the first layer as Query for a second interaction. The output of the second layer is the compressed visual tokens \(\mathbf{F}_{\rm i}^{\rm h}\). 
\end{itemize}

Due to the significant difference between the feature space of \(\mathbf{F}_{\rm i}^{\rm h}\) and the original VLM visual features, we design a dedicated connector \({\rm C}^{\rm h}\) for \(\mathbf{F}_{\rm i}^{\rm h}\).
The entire process can be formalized as:
\begin{equation}
\begin{split}
\mathbf{F}^{\rm L1}&=
{\rm BN}^{\rm L1}({\rm A}_{\rm cross}^{\rm L1}(\mathbf{L}_{\rm h}, {\rm CAT}(\mathbf{F}_{\rm t}^{\rm g}, \mathbf{L}_{\rm h}))),\\
\mathbf{F}_{\rm i}^{\rm h}&={\rm C}^{\rm h}(
{\rm BN}^{\rm L2}({\rm A}_{\rm cross}^{\rm L2}(\mathbf{F}^{\rm L1}, {\rm CAT}(\mathbf{F}_{\rm i}^{\rm h}, \mathbf{F}^{\rm L1})))),\\
\end{split}
\end{equation}
where \({\rm L}{i}\) denotes the \(i^{th}\) layer within DP module, \(\mathbf{F}^{\rm L1}\) is the output of first layer and \({\rm CAT}\) represents the concatenation operation. \({\rm BN}^{{\rm L}i} (i \in (1, 2))\) denotes Bind Network which can be formalized as:
\begin{equation}
\rm{BN}(\mathbf{X})=\mathbf{X}+\left(\mathbf{X} \rm{W_2}^{\rm up} \cdot \operatorname{SiLU}\left(\mathbf{X} \rm{W_1}^{\rm up}\right)\right) \rm{W_3}^{\rm down}.
\end{equation}

\subsection{Overall}
At this point, with the guidance of input text, we have obtained the optimized visual features \(\mathbf{F}_{\rm i}^*\), as well as the detail perceiver tokens \(\mathbf{F}_{\rm i}^{\rm h}\). We then concatenate the features obtained from the original VLM connector \({\rm C}\) with \(\mathbf{F}_{\rm i}^{\rm h}\), which together form the final visual tokens \(\mathbf{T}_{\rm v}^{\rm fin}\) input for the VLM. The final prediction process of VLM can be represented as follows:
\begin{equation}
\begin{split}
\mathbf{T}_{\rm v}^{\rm fin}&={\rm CAT}({\rm C}(\mathbf{F}_{\rm i}^*), \mathbf{F}_{\rm i}^{\rm h}),\\
p\left(\mathbf{T}_{\rm a} \mid \mathbf{T}_{\rm v}^{\rm fin}, \mathbf{T}_{\rm t}\right)&=\prod_{i=1}^L p_{\theta}\left(t_i \mid \mathbf{T}_{\rm v}^{\rm fin}, \mathbf{T}_{{\rm t},<i}, \mathbf{T}_{{\rm a},<i}\right).
\end{split}
\end{equation}

\begin{table*}[t]
\centering
\begin{tabular}{c@{\,}c@{\,}c@{\,}cc@{\,}c@{\,}c@{\,}c@{\,}c@{\,}c@{\,}c@{\,}c@{\,}c@{\,}c}
\hline
Method    & LM         & VE        & \multicolumn{1}{c|}{PT + IT}   & MMB           & MMS           & MMMU          & MV            & OCRB          & AI2D          & HB            & LB            & SQA           & MME           \\ \hline
\multicolumn{14}{c}{\textit{Performance comparison against the baseline}}                                                                                                                                                           \\ \hline
LLaVA-1.5 & Vicuna-7B  & CLIP-L    & \multicolumn{1}{c|}{0.5M+0.6M} & 59.1          & 33.1          & 35.7          & 25.6          & 31.8          & 55.5          & 27.6          & 61.8          & 69.2          & \textbf{1808} \\
TG-LLaVA  & Vicuna-7B  & CLIP-L    & \multicolumn{1}{c|}{0.5M+0.6M} & \textbf{61.3} & \textbf{35.5} & \textbf{38.1} & \textbf{26.7} & \textbf{32.6} & \textbf{56.9} & \textbf{29.2} & \textbf{65.0} & \textbf{70.6} & 1779          \\ \hline
LLaVA-1.5 & Vicuna-13B & CLIP-L    & \multicolumn{1}{c|}{0.5M+0.6M} & 64.0          & 34.3          & 37.0          & \textbf{27.7} & 33.7          & \textbf{61.1} & 24.5          & 66.1          & 72.6          & 1781          \\
TG-LLaVA  & Vicuna-13B & CLIP-L    & \multicolumn{1}{c|}{0.5M+0.6M} & \textbf{65.3} & \textbf{35.9} & \textbf{39.6} & 27.6          & \textbf{34.4} & 61.0          & \textbf{25.9} & \textbf{67.8} & \textbf{72.7} & \textbf{1858} \\ \hline
\hline
\multicolumn{14}{c}{\textit{Expanding to larger training datasets}}                                                                                                                                                                 \\ \hline
LLaVA-1.5 & Vicuna-7B  & CLIP-L    & \multicolumn{1}{c|}{1.2M+1.5M} & 62.8          & 39.0          & 35.2          & \textbf{32.6} & 37.3          & 69.8          & 25.4          & \textbf{60.7} & 70.5          & 1810          \\
TG-LLaVA  & Vicuna-7B  & CLIP-L    & \multicolumn{1}{c|}{1.2M+1.5M} & \textbf{63.5} & \textbf{39.4} & \textbf{37.2} & 32.4          & \textbf{37.9} & \textbf{70.0} & \textbf{27.8} & 59.9          & \textbf{70.9} & \textbf{1840} \\ \hline
\hline
\multicolumn{14}{c}{\textit{Expanding to robust visual encoder}}                                                                                                                                                                    \\ \hline
LLaVA-1.5 & Vicuna-7B  & SigLIP-SO & \multicolumn{1}{c|}{0.5M+0.6M} & 62.8          & 34.9          & 38.6          & 27.0          & 36.3          & \textbf{59.3} & 28.1          & 66.8          & \textbf{70.6} & 1764          \\
TG-LLaVA  & Vicuna-7B  & SigLIP-SO & \multicolumn{1}{c|}{0.5M+0.6M} & \textbf{63.1} & \textbf{37.7} & \textbf{38.9} & \textbf{27.7} & \textbf{37.3} & 58.4          & \textbf{28.5} & \textbf{67.9} & 70.0          & \textbf{1803} \\ \hline
\hline
\multicolumn{14}{c}{\textit{Expanding to other LLMs}}                                                                                                                                                                               \\ \hline
LLaVA-1.5 & Llama3-8B  & CLIP-L    & \multicolumn{1}{c|}{0.5M+0.6M} & \textbf{66.7} & 38.5          & 40.7          & 26.7          & \textbf{33.4} & \textbf{61.8} & 27.4          & 64.3          & 74.8          & 1789          \\
TG-LLaVA  & Llama3-8B  & CLIP-L    & \multicolumn{1}{c|}{0.5M+0.6M} & 65.2          & \textbf{40.5} & \textbf{41.0} & \textbf{28.6} & 32.8          & 60.2          & \textbf{29.2} & \textbf{65.6} & \textbf{75.9} & \textbf{1801} \\ \hline
LLaVA-1.5 & Qwen2-7B   & CLIP-L    & \multicolumn{1}{c|}{0.5M+0.6M} & 70.9          & 42.1          & 43.6          & \textbf{32.2} & \textbf{33.6} & \textbf{65.3} & 28.3          & 65.9          & 74.2          & 1849          \\
TG-LLaVA  & Qwen2-7B   & CLIP-L    & \multicolumn{1}{c|}{0.5M+0.6M} & \textbf{71.2} & \textbf{43.5} & \textbf{44.7} & 31.3          & 33.4          & 64.6          & \textbf{29.2} & \textbf{66.3} & \textbf{75.1} & \textbf{1941} \\ \hline
\end{tabular}
\caption{Performance comparison between various baselines and TG-LLaVA. The results of the first and the third line are sourced from the official OpenCompass publicly available leaderboard~\cite{duan2024vlmevalkit}, while the remaining results are derived from our own replication. The best results are \textbf{bold}. LM, VE, PT and IT denote Language Model, Vision Encoder, pre-training data and instruction fine-tuning data, respectively. 
}
\label{compare_baseline}
\end{table*}

\section{Experiment}
In this section, we first present the detailed experimental setup of our study. We then enumerate the improvements brought by our proposed TG-LLaVA over the baseline across multiple evaluation metrics, and compare our method with several state-of-the-art (SoTA) approaches under various configurations. 
Specifically, we visualize the attention map to demonstrate the efficacy of proposed TG-LLaVA. 
Finally, we conduct ablation studies and provide an analysis of the results.

\subsection{Experimental settings}
\subsubsection{Implementation Details}
We implement the proposed improvement strategy on top of LLaVA-1.5~\cite{liu2024improvedbaselinesvisualinstruction}, whose general applicability in the VLM field facilitates the validation of our method's versatility. 
Specifically, we maintain consistency with LLaVA-1.5 by employing CLIP-ViT-L/14-336px as the visual encoder. To further validate the generalizability of our proposed method, we also incorporate SigLIP-SO400m-patch14-384, another leading choice, for comparative analysis. In terms of LLM, we compare our method against the baseline using Vicuna-7/13B and extend our approach to Llama3-8B~\cite{llama3} and Qwen2-7B~\cite{yang2024qwen2}, thereby demonstrating the versatility of our method. For training configurations, we adhere strictly to the settings outlined in the original LLaVA-1.5 paper to ensure fairness, with learning rates of 1e-3 and 2e-5 for pre-training and instruction fine-tuning phases, respectively, and maintaining batch sizes of 256 and 128. DP module introduces 64 additional visual tokens. The training process for TG-LLaVA utilizes the PyTorch framework and employs 8 H100-80G GPUs. 

\subsubsection{Datasets}
Focusing on proposing a novel optimization method for the VLM framework, we do not incorporate any additional data beyond the LLaVA-1.5 open-source dataset~\cite{liu2024improvedbaselinesvisualinstruction}, which includes 558K image captions for pre-training and 665K conversations for instruction tuning. We also apply our proposed method to the Mini-Gemini dataset~\cite{reid2024gemini}, which consists of 1.2M + 1.5M data, to further highlight the superiority of our approach. For evaluation, we conduct extensive experiments and report results on widely-adopted VLM benchmarks using the VLMEvalKit~\cite{duan2024vlmevalkit} platform to provide robust and comprehensive performance validation for the proposed TG-LLaVA. The evaluation datasets include: MMBench (MMB)~\cite{liu2023mmbench}, MMS (MMStar)~\cite{chen2024we}, MMMU~\cite{yue2024mmmu}, MV (MathVista)~\cite{lu2023mathvista}, OCRB (OCRBench)~\cite{liu2023hidden}, AI2D~\cite{hiippala2021ai2d}, HB (HallusionBench)\cite{guan2024hallusionbench}, LB (LLaVABench)~\cite{liu2024visual}, SQA (ScienceQA)~\cite{saikh2022scienceqa}, and MME~\cite{fu2024mmecomprehensiveevaluationbenchmark}.

\subsection{Genuine Improvement Over the Baseline}
In Table \ref{compare_baseline}, we present the performance improvements of the proposed method across various configurations compared to the baseline. According to the experimental results, we can draw several phenomenons:
\begin{itemize}
    \item The proposed text-guided strategy demonstrates substantial improvements over the baseline.  
    Compared with the original LLaVA-1.5, TG-LLaVA achieve much better performance. As shown in the first four rows. our method leads on the majority of evaluation datasets. It is noteworthy that TG-LLaVA demonstrates an average improvement of 1.5\% over the original LLaVA-1.5 across ten datasets when using Vicuna-7B, highlighting the method's significant value. When juxtaposed with the baseline LLaVA-1.5 Vicuna-7B model, we enhance performance metrics by +2.2\% on MMBench, +2.4\% on both MMStar and MMMU, and +3.2\% on LLaVABenchs, respectively. For LLaVA-1.5 with Vicuna-13B, we also achieve an average performance improvement of 1\%. Specifically, we see a +1.6\% gain on MMStar, a +2.0\% gain on MMMU, and a +3.2\% gain on MME. These impressive results further validate the contribution of the proposed TG-LLaVA architecture to visual feature optimization, highlighting the favorable impact of our method. 
    \item The proposed TG-FOM and TG-DP modules can be universally applied as a modular plugnin to mainstream VLM frameworks. As shown in the rest part of Table \ref{compare_baseline}. we further validate the versatility of our proposed method under various settings. We replace CLIP with SigLIP and substitute Vicuna with Llama3 and Qwen2 on top of the original LLaVA-1.5 framework. We compare these settings with our method as the baseline. The results in Table \ref{compare_baseline} confirm that our method continues to maintain a leading advantage across most datasets, demonstrating that the proposed TG-LLaVA exhibits excellent generalizability and possesses strong potential for adaptation to a wide range of VLM architectures.
\end{itemize}

\begin{table}[t]
\centering
\setlength\tabcolsep{2pt}
\begin{tabular}{cccccc}
\hline
Method          &   Source     & MME           & MMB           & MMVet         & GQA           \\ 
\hline
LLaVA-1.5       &    NeurIPS 23  &  1531          & 67.7          & 35.4          & 63.3          \\
\hline
Seeing the image &  Arxiv 2405  & 1567          & -             & -             & -             \\
TokenPacker      &  Arxiv 2407  &-          & 68.0          & 34.5          & 62.5          \\
DenseConnector\rlap{\(^\star\)}   &  Arxiv 2405  & 1540          & 70.0         & -             & -             \\
\hline
TG-LLaVA         &  -            & \textbf{1603} & \textbf{70.2} & \textbf{36.6} & \textbf{63.4}
\\ \hline
\end{tabular}
\caption{Performance comparison with contemporaneous methods. \(\star\) denotes results obtained with official code reproductions. Note: MME metric here considers only the Perception part.}
\label{compare_recent}
\end{table}

\begin{table*}[t]
\centering
\begin{tabular}{c@{\,}c@{\,}c@{\,}cc@{\,}c@{\,}c@{\,}c@{\,}c@{\,}c@{\,}c@{\,}c@{\,}c@{\,}c}
\hline
Method                          & LLM                               & VE                               & \multicolumn{1}{c|}{PT + IT}                                               & MMB                                  & MMS                                  & MMMU                                 & MV                                   & OCRB                        & AI2D                                 & HB                          & LB                                   & SQA                                  & MME                                  \\ \hline
MiniGPT4                        & Vicuna-7B                         & EVA-G                            & \multicolumn{1}{c|}{5M+3.5K}                          & 20.8                                 & 16.3                                 & 23.6                                 & 20.4                                 & 17.2                        & 28.4                                 & \underline{31.9}                        & 45.1                                 & 39.6                                 & 1047                                 \\
Qwen-VL                          & Qwen-7B                           & ViT-G/16                         & \multicolumn{1}{c|}{1.4B+50M}                         & 32.9                                 & 32.5                                 & 29.6                                 & 15.5                                 & 12.7                        & 57.7                                 & 29.9             & 12.9                                 & 61.1                                 & 483                                  \\
VisualGLM                       & ChatGLM-6B                        & EVA-CLIP                         & \multicolumn{1}{c|}{330M}                             & 35.7                                 & 25.9                                 & 29.9                                 & 21.9                                 & 17.0                        & 41.2                                 & 25.0                        & 37.3                                 & 56.1                                 & 738                                  \\
PandaGPT                        & Vicuna-13B                        & IB-H                             & \multicolumn{1}{c|}{160K}                             & 34.5                                 & 25.6                                 & 32.9                                 & 25.0                                 & 26.9                        & 48.3                                 & 21.6                        & 57.2                                 & 61.8                                 & 1076                                 \\
mPLUG-Owl2                      & Llama 2-7B                        & CLIP-L                           & \multicolumn{1}{c|}{348M+1.2M}                        & 60.8                                 & 34.8                                 & 34.7                                 & 25.4                                 & 25.5                        & 55.7                                 & 29.4                        & 59.9                                 & 69.5                                 & 1786                                 \\
Emu2-chat                       & Llama-33B                         & EVA-CLIP                         & \multicolumn{1}{c|}{-}                                & 52.8                                 & \underline{40.7}                                 & 35.0                                 & 30.7                                 & \textbf{43.6}                & 49.7                                 & 29.5                        & 56.4                                 & 68.2                                 & 1678                                 \\
Yi-VL                           & Yi-6B                             & CLIP-L                           & \multicolumn{1}{c|}{100M+26M}                         & 64.2                                 & 33.7                                 & 40.3                                 & 29.7                                 & 29.0                         & 59.8                                 & \textbf{36.0}                        & 51.9                                 & 72.6                                 & \underline{1915}                                 \\
ShareGPT-4V                     & Vicuna-7B                         & CLIP-L                           & \multicolumn{1}{c|}{1.2M+0.7M}                        & 61.6                                 & 35.7                                 & 37.2                                 & 26.5                                 & 37.1                         & 58.0                                 & 28.6                        & \underline{66.9}                                 & 69.5                                 & 1914                                 \\ \hline \hline
TG-LLaVA & Vicuna-7B  & SigLIP-SO & \multicolumn{1}{c|}{0.5M+0.6M} & 63.1 & 37.7 & 38.9 & 27.7 & 37.3 & 58.4 & 28.5 & \textbf{67.9} & 70.0 & 1803 \\
TG-LLaVA & Vicuna-7B  & CLIP-L    & \multicolumn{1}{c|}{1.2M+1.5M} & 63.5 & 39.4 & 37.2 & \textbf{32.4} & \underline{37.9} & \textbf{70.0} & 27.8 & 59.9 & 70.9 & 1840 \\
TG-LLaVA & Llama3-8B  & CLIP-L    & \multicolumn{1}{c|}{0.5M+0.6M} & \underline{65.2} & 40.5 & \underline{41.0} & 28.6 & 32.8 & 60.2 & 29.2 & 65.6 & \textbf{75.9} & 1801 \\
TG-LLaVA & Qwen2-7B   & CLIP-L    & \multicolumn{1}{c|}{0.5M+0.6M} & \textbf{71.2} & \textbf{43.5} & \textbf{44.7} & \underline{31.3} & 33.4 & \underline{64.6} & 29.2 & 66.3 & \underline{75.1} & \textbf{1941} \\ \hline
\end{tabular}
\caption{Comparison with SoTA methods. The best results are \textbf{bold} and the second-best results are \underline{underlined}. Results of all other methods are obtained from the OpenCompass public leaderboard.}
\label{compare_sota}
\end{table*}

\subsection{Comparison with other LLaVA-based methods}
In Table \ref{compare_recent}, we compare the proposed proposed TG-LLaVA with other concurrent works which also take LLaVA as baseline. The methods we include for comparison are Seeing the Image~\cite{xiao2024seeing}, TokenPacker\cite{li2024tokenpacker}, and DenseConnector\cite{yao2024dense}. Since these comparison methods are relatively new and have not been evaluated on the OpenCompass leaderboard, we employ the evaluation scripts from LLaVA-1.5 to maintain a fair and consistent framework for our comparisons. 

\subsection{Quantitative Comparison with SoTAs}
We further compare our method with several leading approaches. The methods included in the comparison are MiniGPT4~\cite{zhu2023minigpt}, Qwen-VL\cite{bai2023qwenvl}, VisualGLM\cite{glm2024chatglm}, PandaGPT\cite{su2023pandagpt}, mPLUG-Owl2\cite{ye2023mplugowl2revolutionizingmultimodallarge}, Emu2-chat\cite{sun2024generativemultimodalmodelsincontext}, Yi-VL\cite{young2024yi} and ShareGPT-4V\cite{chen2023sharegpt4vimprovinglargemultimodal}. Table \ref{compare_sota} presents the performance comparison across multiple benchmarks.

Remarkably, despite relying solely on settings from LLaVA-1.5, our TG-LLaVA achieves performance that matches or surpasses the benchmarks set by leading SoTA methods, with a comparatively smaller volume of pre-training and instruction fine-tuning data.

\begin{figure}[t]
\centering
\includegraphics[width=0.47\textwidth]{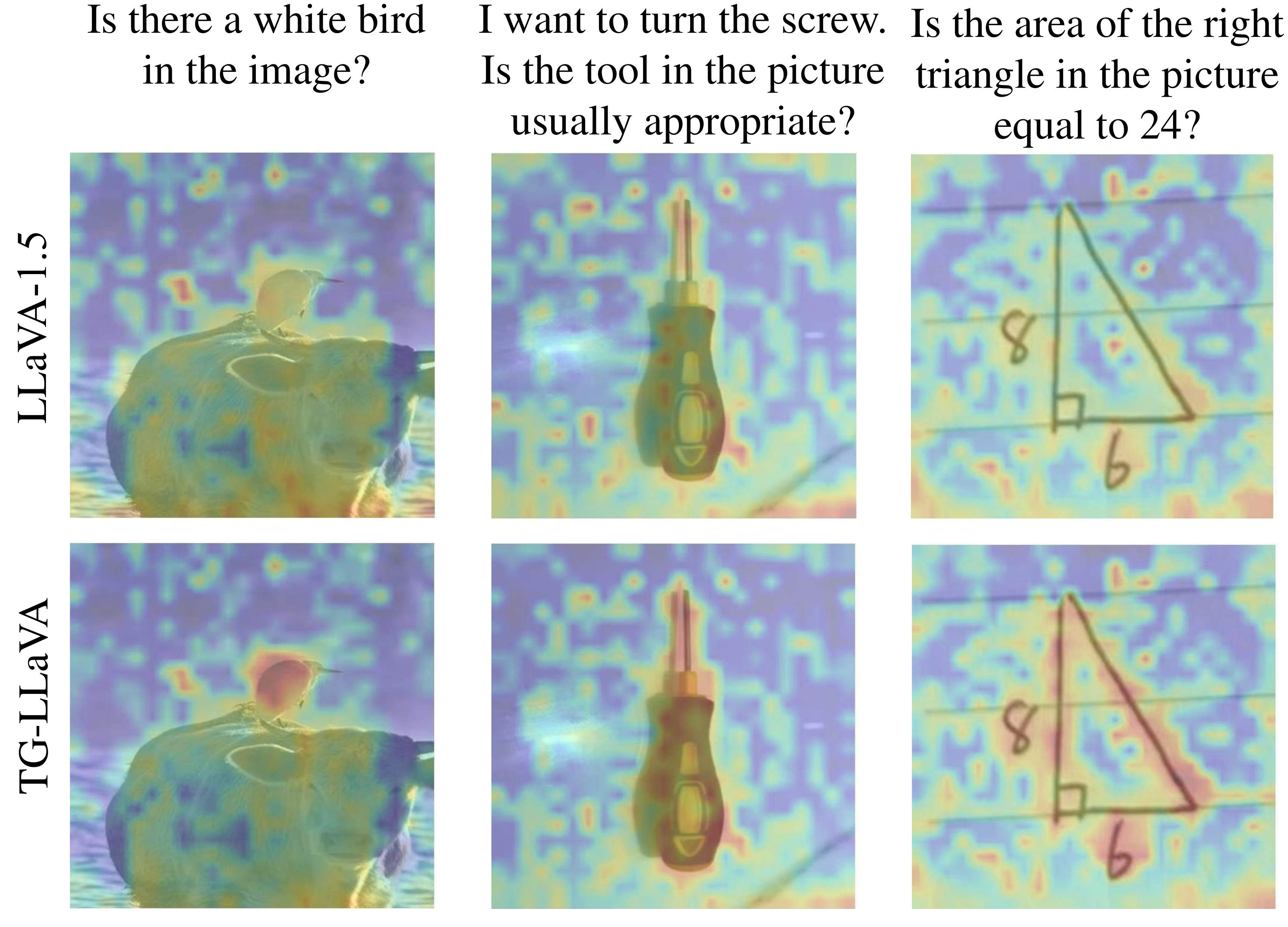} 
\caption{Attention maps for TG-LLaVA versus LLaVA-1.5 using Llama3-8B. }
\label{TG_visual}
\end{figure}

\subsection{Qualitative Analysis via Visualization}
To demonstrate the optimization effect of the proposed method, we visualize the attention maps between the visual features and text instructions in both the baseline model and our model. These visualizations provide insights into how the proposed visual feature optimization module operates. We aggregate the attention scores between image tokens and textual instruction tokens across all layers to compute the results. As shown in Figure \ref{TG_visual}, the TG-LLaVA architecture push the model to focus on regions highlighted by the textual instructions, assigning greater attention weights to them.

\begin{table}[t]
\centering
\setlength\tabcolsep{10pt}
\begin{tabular}{c|cccc}
\hline
    Setting         & MMB   & MV     & AI2D         & SQA \\ \hline
Baseline     & 59.1          & 25.6           & 55.5          & 69.2           \\ \hline
Only FOM     & 60.4          & 26.0             & 55.1          & 67.5           \\
Only DP      & 58.7          & 26.1           & 56.0            & 69.3           \\ 
DP patch 32  & 60.7          & 26.1           & 56.7          & 70.5           \\
DP patch 128 & 60.5          & 26.2           & 56.9          & 69.1           \\
DP patch 256 & 61.3 & 25.9           & 55.4          & 68.5           \\ \hline
Final        & 61.3 & 26.7 & 56.9 & 
70.6 
\\ \hline
\end{tabular}
\caption{Ablation study results on FOM and DP modules, and impact of the additional visual token count introduced in DP module.}
\label{compare_ab}
\end{table}

\subsection{Ablation Studies}
We further conduct in-depth ablation studies to analyze the effectiveness of each component of our approach. Results are listed in Table \ref{compare_ab}. By sequentially introducing the FOM and DP modules, we observe significant improvements in model performance, underscoring the effectiveness of our proposed visual feature optimization algorithm. 
Additionally, we conduct experiments on the number of additional visual tokens introduced by the DP module. The results show that introducing too few tokens yields suboptimal performance gains, while introducing too many tokens can actually harm performance. Therefore, we choose a balanced configuration to achieve optimal performance.


\section{Conclusion}
In this paper, we introduce TG-LLaVA, an innovative VLM optimization technique that guides the vision encoder using text. By emulating human-like purpose-driven logic, we leverage learnable embeddings to analyze text and enhance the vision encoder. Our experiments reveal that TG-LLaVA outperforms similar methods and is adaptable to various frameworks, consistently yielding improvements. This text-guided enhancement of the visual encoder opens up a new pathway for advancing VLMs. For future work, we aim to further refine the visual feature extraction process guided by text to achieve even better performance.

\bibliography{aaai25}

\newpage
\section{Appendix}

\subsection{Experiment Settings}
TG-LLaVA employs the same set of hyperparameters as LLaVA-1.5. It is important to note that both the TG-FOM and TG-DP modules in the proposed architecture are trained from scratch without any form of pre-trained parameters. Table \ref{setting} summarizes the training hyperparameters for both the first phase of visual-language alignment pre-training and the second phase of visual instruction tuning for TG-LLaVA.


\subsection{How to guide the vision encoder}
As discussed in main paper, our objective is to explore how to endow VLMs with purpose-driven reasoning capabilities akin to human logic to achieve superior model performance. The key lies in how to utilize textual instructions to guide the VLMs, enabling a "focus" on the visual modality.

In SAM-Adapter~\cite{chen2023samfailssegmentanything}, the authors use low-rank parameter matrices to handle task-specific features (such as high-frequency components obtained from Fast Fourier transforms of images to be segmented) and inject these features layer by layer into SAM’s~\cite{kirillov2023segment} visual encoder (a Vision Transformer structure). This approach enables the transition of SAM from general domains to medical image segmentation. Inspired by this, we attempt a similar approach by using TG-FOM module to inject textual instruction information into the shallow layers of the VLM’s visual encoder. As shown in Table \ref{insert}, we experiment with various strategies: inserting the output of TG-FOM module at the head, middle, and tail of the visual encoder, as well as combinations of these positions. 
We use MME dataset metric as the evaluation standard and CLIP-L with 24 layers as an example, insertion at the head refers to adding features between the Patch Embedding and the first layer of the Transformer Encoder. Insertion at the middle occurs between the 12th and 13th layers, while insertion at the tail refers to applying it at the output.
However, the results are disappointing. Except for the final approach, which inserts the module at the encoder's output, the other shallow-layer modifications lead to noticeable degradation in performance. We consider this is due to the fact that in the shallow layers of the visual encoder, the visual features are not yet aligned with the textual features in a unified space. Thus, even with a zero-initialized linear layer as a buffer, the outcome still results in negative gains. Further exploration is required to determine how to effectively integrate textual instruction information during the visual encoding process.

\subsection{Model Zoo}
We further explore the upper limits of TG-LLaVA by fine-tuning on Qwen2-7B model with a larger dataset, utilizing the training data introduced by Ovis~\cite{lu2024ovisstructuralembeddingalignment}. The Ovis training data comprises open-source datasets and a small portion of internal datasets, categorized into three types: visual captions, visual descriptions, and multimodal instructions, used for the three-stage training process.
We use the 10M visual captions subset as the training data for the second stage in training, with the results displayed alongside other variants in Table~\ref{ovis}. As shown, underpinned by a robust dataset, TG-LLaVA demonstrates a remarkable average increase of 8\% across ten datasets. Notably, it achieves a 20.2\% improvement on MathVista and a 12.2\% increase on ScienceQA, underscoring TG-LLaVA's significant potential and the crucial role that high-quality data plays in enhancing VLM performance.

\begin{table}[t]
\centering
\begin{tabular}{c|cc}
\hline
Hyperparameter  & Pretraining & Instruction Tuning \\ \hline
batch size      & 256         & 128                \\
learning rate   & 1e-3        & 2e-5               \\
schedule        & \multicolumn{2}{c}{cosine decay} \\
warmup ratio    & \multicolumn{2}{c}{0.03}         \\
weight decay    & \multicolumn{2}{c}{0}            \\
epoch           & \multicolumn{2}{c}{1}            \\
optimizer       & \multicolumn{2}{c}{AdamW}        \\
DeepSpeed stage & 2           & 3                  \\ \hline
\end{tabular}
\caption{Hyperparameters of TG-LLaVA.}
\label{setting}
\end{table}

\begin{table}[t]
\centering
\begin{tabular}{c|c}
\hline
Insert Strategy      & MME  \\ \hline
head                 & 711  \\
middle               & 694  \\
middle + tail        & 682  \\
head + middle + tail & 687  \\
tail                 & 1603 \\ \hline
\end{tabular}
\caption{Effects evaluation of inserting textual instruction features at different positions.}
\label{insert}
\end{table}

\begin{table*}[t]
\centering
\begin{tabular}{c@{\,}c@{\,}c@{\,}cc@{\,}c@{\,}c@{\,}c@{\,}c@{\,}c@{\,}c@{\,}c@{\,}c@{\,}c}
\hline
Method   & LM       & VE     & \multicolumn{1}{c|}{PT + IT}  & MMB  & MMS  & MMMU & MV   & OCRB & AI2D & HB   & LB   & SQA  & MME  \\ \hline
TG-LLaVA  & Vicuna-7B  & CLIP-L    & \multicolumn{1}{c|}{0.5M+0.6M} & 61.3 & 35.5 & 38.1 & 26.7 & 32.6 & 56.9 & 29.2 & 65.0 & 70.6 & 1779          \\ 
TG-LLaVA  & Vicuna-13B & CLIP-L    & \multicolumn{1}{c|}{0.5M+0.6M} & 65.3 & 35.9 & 39.6 & 27.6          & 34.4 & 61.0          & 25.9 & 67.8 & 72.7 & 1858 \\ 
TG-LLaVA  & Vicuna-7B  & CLIP-L    & \multicolumn{1}{c|}{1.2M+1.5M} & 63.5 & 39.4 & 37.2 & 32.4          & 37.9 & 70.0 & 27.8 & 59.9          & 70.9 & 1840 \\ 
TG-LLaVA  & Vicuna-7B  & SigLIP-SO & \multicolumn{1}{c|}{0.5M+0.6M} & 63.1 & 37.7 & 38.9 & 27.7 & 37.3 & 58.4          & 28.5 & 67.9 & 70.0          & 1803 \\ 
TG-LLaVA  & Llama3-8B  & CLIP-L    & \multicolumn{1}{c|}{0.5M+0.6M} & 65.2          & 40.5 & 41.0 & 28.6 & 32.8          & 60.2          & 29.2 & 65.6 & 75.9 & 1801 \\ 
TG-LLaVA  & Qwen2-7B   & CLIP-L    & \multicolumn{1}{c|}{0.5M+0.6M} & 71.2 & 43.5 & 44.7 & 31.3          & 33.4          & 64.6          & 29.2 & 66.3 & 75.1 & 1941 \\ 
TG-LLaVA & Qwen2-7B & CLIP-L & \multicolumn{1}{c|}{0.5M+10M} & \textbf{74.2} & \textbf{51.5} & \textbf{46.4} & \textbf{52.6} & \textbf{48.9} & \textbf{78.5} & \textbf{37.2} & \textbf{71.7} & \textbf{88.1} & \textbf{1993} \\ \hline
\end{tabular}
\caption{More comprehensive evaluation results of TG-LLaVA.}
\label{ovis}
\end{table*}

\begin{table}[t]
\centering
\begin{tabular}{c|cc|cc}
\hline
\multirow{2}{*}{} & \multicolumn{2}{c|}{Vicuna-7B} & \multicolumn{2}{c}{Vicuna-13B} \\ \cline{2-5} 
                  & PT             & IT            & PT            & IT             \\ \hline
LLaVA-1.5         & 2.0            & 6.7           & 2.8           & 10.8           \\
TG-LLaVA          & 2.2            & 7.4           & 3.1           & 11.6           \\ \hline
\end{tabular}
\caption{Comparison of two-Stage training time consumption under different settings.}
\label{time}
\end{table}

\subsection{Training Cost Analysis}
The proposed TG-LLaVA architecture involves multiple attention computations and image patch segmentation, leading to increased computational resource consumption during training. We record the training time for both the baseline LLaVA-1.5 and TG-LLaVA across two stages, with experiments conducted on 8 H100 GPUs. As shown in Table~\ref{time}, TG-LLaVA incurs approximately 10\% more time compared to the baseline. The additional trainable parameters introduced by TG-LLaVA amount to 0.18B compared to LLaVA-1.5 with Vicuna-13B. Currently, the implementation involves processing multiple detailed patches sequentially through the visual encoder, but optimizing this to a batch format could further reduce training delays.

\subsection{Qualitative Results}
\begin{figure*}[th]
\centering
\includegraphics[width=0.95\textwidth]{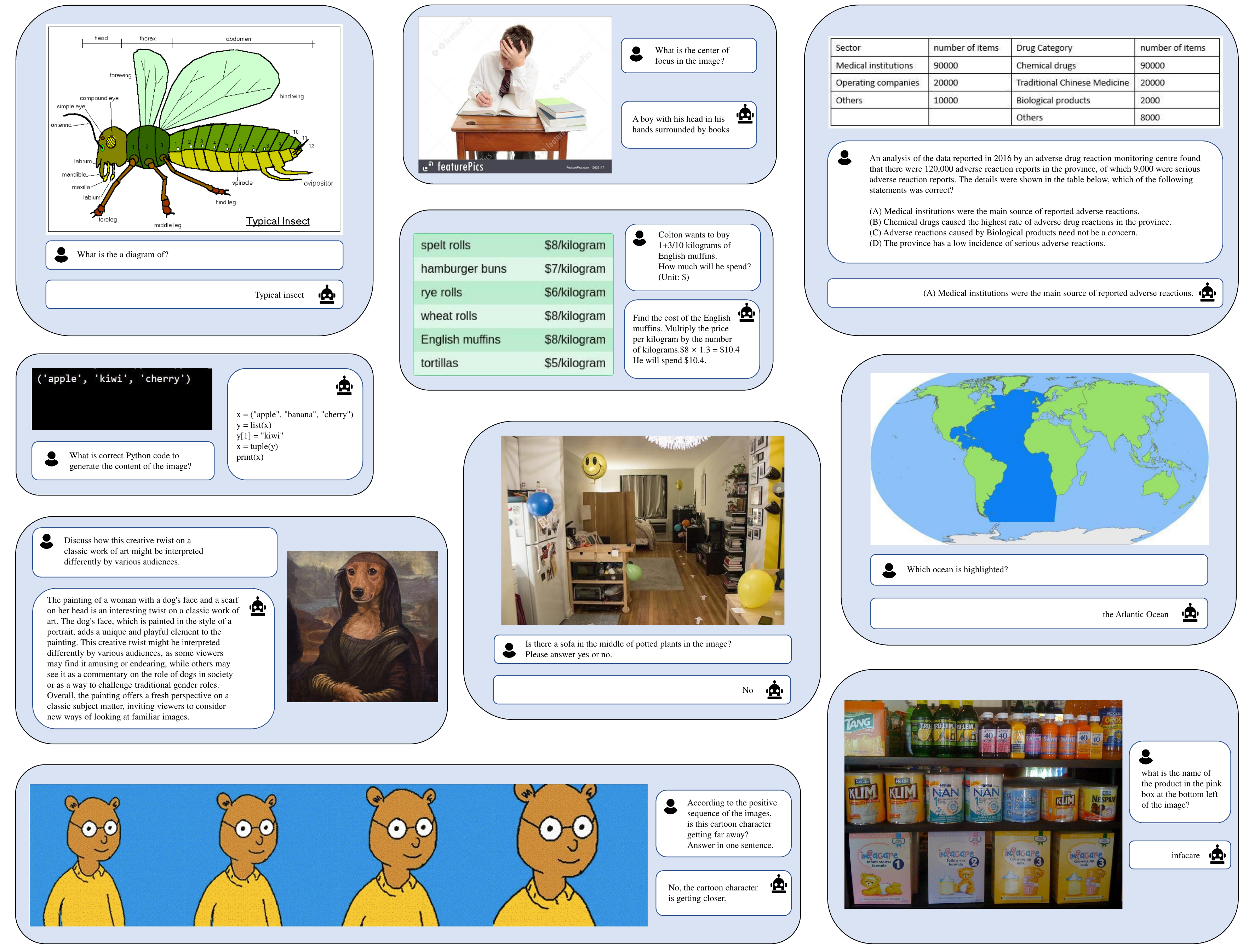} 
\caption{Inference examples of TG-LLaVA on the ten datasets involved in the experimental section.}
\label{example}
\end{figure*}

We validate the inference capabilities of TG-LLaVA across various tasks involving understanding and reasoning to demonstrate its effectiveness in practical scenarios. As shown in Figure~\ref{example}, TG-LLaVA handles complex visual-language reasoning problems in diverse contexts, including image perception and comprehension, mathematics and computation, OCR recognition, and other tasks requiring relevant prior knowledge.

\subsection{Limitation and Future Work}
To achieve text-guided visual feature optimization, TG-LLaVA inevitably introduces additional computational resource consumption, including extended training time and increased parameter count, beyond the original LLaVA-1.5. A viable improvement approach is to compress the number of visual tokens, such as TokenPacker~\cite{li2024tokenpacker}. Compared to the total parameters of the original VLM, the increased parameter count is relatively small, and we believe that these trade-offs are reasonable without increasing the training data. In addition, our method can be used as a plug-in module to further improve the performance of other VLM models with similar architectures. Future work will explore methods to further reduce computational resource consumption while refining text-guided visual enhancement and improving guidance strategies.

\end{document}